# Real time expert system for anomaly detection of aerators based on computer vision technology and existing surveillance cameras

Yeqi Liu[1,2,3], Yingyi Chen[1,2,3,*], Huihui Yu[1,2,3], Xiaomin Fang[1,2,3], Chuanyang Gong[1,2,3]

*1. College of Information and Electrical Engineering, China Agricultural University, Beijing 100083, P.R. China;*
*2. Key Laboratory of Agricultural Information Acquisition Technology, Ministry of Agriculture, Beijing 100083, P.R. China;*
*3. Beijing Engineering and Technology Research Center for Internet of Things in Agriculture, Beijing 100083, P.R. China.*
*\* Corresponding author (E-mail: chyingyi@126.com).*

**Abstract**

Aerators are essential and crucial auxiliary devices in intensive culture, especially in industrial culture in China. However, due to overload operation and poor working condition, the aerator will often has abnormal condition but cannot be detected in time, which will lead to rapid death of fish in large areas due to lack of oxygen, and then cause huge economic losses and ecological problems. The traditional methods cannot accurately detect abnormal condition of aerators in time. The manual inspection method is laborious and inefficient, and circuit detection methods often suffer from error detection or short circuit. Therefore, it is necessary to develop a real-time expert system for anomaly detection of aerators. Surveillance cameras are widely used as visual perception modules of the Internet of Things, but they are only used to detect suspicious behavior of human being in aquaculture in China. Using these widely existing surveillance cameras to realize real-time anomaly detection of aerators is a cost-free and easy-to-promote method, which also provides a useful reference and attempt for intelligent agriculture and agricultural expert systems. However, it is difficult to develop such an expert system due to some technical and applied challenges, e.g., illumination, occlusion, complex background, etc.

To tackle these aforementioned challenges, we propose a real-time expert system based on computer vision technology and existing surveillance cameras for anomaly detection of aerators, which consists of two modules, i.e., object region detection and working state detection. First, it is difficult to detect the working state for some small object regions in whole images, and the time complexity of global feature comparison is also high, so we present an object region detection method based on the region proposal idea. Moreover, we propose a novel algorithm called reference frame Kanade-Lucas-Tomasi (RF-KLT) algorithm for motion feature extraction in fixed regions. Then, we present a dimension reduction method of time series for establishing a feature dataset with obvious boundaries between classes. Finally, we use machine learning algorithms to build the feature classifier. The experimental results show that the accuracy for detecting object region and working state of aerators is 100% and 99.9% respectively, and the detection speed is 77-333 frames per second (FPS) according to the different types of surveillance cameras. The proposed expert system can realize real-time, robust and cost-free anomaly detection of aerators in both the actual video dataset and the augmented video dataset.

## 1. Introduction

With the development of informatization and intelligentization, aquaculture is also increasingly tending to intensive culture and even industrial culture. Automated monitoring and control is critical to the development of this direction. Dissolved oxygen is one of the key parameters of water quality in aquaculture which must be detected and regulated, because maintaining the stability of dissolved oxygen and related parameters is of great significance for ensuring the safe and rapid growth of aquatic animals and plants (Bardon-Albaret & Saillant, 2016; Chen, Xu, Yu, Zhen, & Li, 2016; Solstorm et al., 2018). The impeller aerator is widely used in China to ensure the stability of dissolved oxygen and related parameters due to their low-cost and high-efficiency features. Hence, real-time anomaly detection of aerators is crucial to avoid economic losses and ecological problems caused by rapid death of fish in large areas due to lack of oxygen.

Traditionally, the detection of abnormal condition of aerators mainly depends on manual inspection, and some researchers also try to use circuit detection method (Ma, Zhao, Wang, Chen, & Li, 2015). However, the former method is labor-intensive and inefficient, and the latter often suffers from error detection or short circuit, which also cannot detect equipment problems caused by mechanical failure, e.g., the machine is stuck due to lubrication or foreign matter. Besides, the circuit maintenance requires professionals to operate, which is dangerous and difficult for ordinary workers. These methods cannot find abnormal condition timely and accurately.

Recently, computer vision technology has been one of the most rapidly developing and mature subfields in artificial intelligence. It has been widely used in various fields to achieve industrialization, such as face recognition (Lu & Tang, 2014), automatic driving (Janai, Güney, Behl, & Geiger, 2017), plant identification (Wäldchen & Mäder, 2018), etc. In short, computer vision technology generally has the characteristics of no human contact, high efficiency and high precision. Furthermore, surveillance cameras are widely used as the visual perception module of the Internet of Things in a variety of applications, including aquaculture. The video-based method further extends the application scope of computer vision technology, which can not only use a single image processing technology, but also provide more information in the temporal dimension of the inter-frame connection. Computer vision technology based on video images is used to detect the change of motion state in traffic monitoring and security field (Jun, Aggarwal, & Gokmen, 2016; Murugan, Jacintha, & Shifani, 2017).

Therefore, the use of existing surveillance cameras and computer vision technology can achieve real-time anomaly detection of aerator without any extra cost, which also can reduce labor intensity. Meanwhile, this methodology can realize real-time warning and recording of abnormal condition, providing useful reference and attempt for intelligent agriculture and agricultural expert systems. However, the methodology for robust object region detection and real-time working state detection of aerators using computer vision technology and existing surveillance cameras faces three major challenges, which have restricted the development of the methodology and application.

1) In the actual environment, the height of the existing surveillance camera and its position relative to the target object are different, which results in a large difference in the size and shape of the object region in videos, e.g., small object, long strip region, etc.

2) There are many interference factors caused by different weather, different time, and different scenes, etc., such as occlusion, brightness changes, camera jitter, background interference including ripple, suspended matter, pedestrian interference, etc.

3) Some system performance required by the expert system, e.g., stable, robust, fast, low-cost, easy to operate, etc.

To address these aforementioned challenges, this paper proposes the RF-KLT algorithm for motion feature extraction and presents a dimension reduction method for feature dataset establishment. The aim of this research is to provide a real-time and automatic expert system for anomaly detection of aerators with existing surveillance cameras. The study is conducted in two modules: (1) object region detection. For this purpose, the procedure of maximum contour region detection, candidate region detection, and object region detection is designed; (2) working state detection. The image feature of each frame is extracted by RF-KLT algorithm, and then the time series is converted into a two-dimensional feature dataset. Finally, the feature classification model established by SVM algorithm is used to detect the working state. The comparison between multiple foreground detection algorithms and machine learning algorithms shows that the proposed methods perform well in both modules.

*1.1. Contributions*

The main contributions of our study are three-fold**:**

1) RF-KLT algorithm for motion feature extraction. To the best of our knowledge, we first propose a novel algorithm called RF-KLT algorithm combining with the reference idea and KLT algorithm (Shi, 1994). Specifically, we assume that all motions are changes relative to a fixed reference frame, extending the conditions under which the KLT algorithm requires continuous inter-frame motion. We apply the RF-KLT algorithm to the object region detection and motion feature extraction of aerators, which proves that this algorithm can judge whether the current frame has a specific motion pattern relative to the reference frame.

2) Object region detection method. Based on the latest region proposal idea (Girshick, Donahue, Darrell, & Malik, 2014) and the proposed RF-KLT algorithm, we present a small object region detection method consisting of three steps, i.e., maximum contour region detection, candidate region detection, and object region detection. In our application scenario, the problem of small object region caused by long distance and high position of surveillance cameras is solved.

3) Time series dimension reduction method for feature dataset construction. We introduce a time series dimension reduction method for constructing feature dataset, i.e., we construct a two-dimensional dataset by using the average value of the window time series as another dimension based on the numerical distribution features of time series, thereby training a two-class classifier for detecting the working state of aerators.

4) Expert system in practical applications. We present an expert system for anomaly detection of aerators, which consists of object region detection and working state detection. The object region detection method refers to the procedure of selecting target region from candidate regions, and the working state detection method includes motion feature extraction, feature dataset construction, and classifier training. In the actual application scenarios, as well as the augmented videos with artificial noise and interference in multiple ways, the proposed expert system can realize real-time, stable and accurate anomaly detection of aerators.

*1.2. Paper organization*

The rest of this paper is structured as follows. Section 2 presents an overall review of traditional and state-of-the-art related works. In Section 3, we first introduce the raw video data and the artificially augmented video data. Then, we organize the introduction of all methods according to the technical flow of object region detection and work state detection. Notably, the RF-KLT algorithm and the dataset construction method of time series are both presented in Section 3.2.3. In Section 4, we show the comparison, discussion and experimental results of the proposed expert system. Finally, Section 5 shows the conclusions and the future work of this study.

**2. Related work**

Because of the three major challenges we have summarized in Introduction, there are few existing studies based on computer vision and surveillance cameras for aerator-related detection. He et al. (Jinhui He, 2015) has made some attempts in this application scenario. They used Harris corner features based on the computer vision technology for working state detection between adjacent frames with the detection time of 4 FPS, which shows the real-time performance is not well. Moreover, they presented artificially selected object region, which inevitably affects the extraction of feature points. Additionally, the inter-class distance of the dataset established using the statistical values of the motion feature distribution between adjacent frames is not obvious. By contrast, we extract the motion features based on RF-KLT algorithm to select the object region in the complex background, and propose a dataset construction method with obvious inter-class distance of the feature dataset.

**Expert systems based on computer vision technology and surveillance cameras.** Expert systems based on computer vision technology and surveillance cameras are widely used in various fields, e.g., traffic detection (Santhosh Kelathodi Kumaran, 2018), pedestrian identification (Khan, Park, & Kyung, 2018), fire detection (Frizzi et al., 2016), etc. Expert systems that use surveillance cameras to detect suspicious behavior have also received attention from researchers (Arroyo, Yebes, Bergasa,

Daza, & Almazán, 2015; Lee, Leong, Lai, Leow, & Yap, 2018). In addition, expert systems based on computer vision technology and video information are used in agriculture for counting (Wadhai, Gohokar, & Khaparde, 2015), tracking (Chuang, Hwang, Ye, Huang, & Williams, 2017), and behavioral analysis of animals (Pasupa, Pantuwong, & Nopparit, 2015). However, most of these methods, especially in agriculture, are based on a particular environment or require the relative position of the monitoring device to the monitored object, which limits the application value of this technology. Obviously, real-time and on-the-ground monitoring combined with existing surveillance cameras and sophisticated computer vision technology is a low-cost, easy-to-promote, and real-time method in many application scenarios.

**Object detection in video.** Object detection methods in video are based on inter-frame features. The video adds time dimension to a single image, which helps to detect the foreground region (Toyama, Krumm, Brumitt, & Meyers, 1999). Andrews et al. (Sobral & Vacavant, 2014) compares 29 video-based background subtraction algorithms implemented in BGSLibrary for object detection. Nevertheless, the classical background modeling algorithms can detect a small foreground region (Zivkovic & Ferdinand, 2006), but they cannot feedback other features of the foreground region, so they are still difficult to further detect the object region from a large number of candidate object regions. Recently, object detection in videos of ImageNet introduce a new challenge for this domain, i.e. VID, and all state-of-the-art methods are based on deep neural networks, e.g., RCNN (Girshick et al., 2014), YOLO (Redmon, Divvala, Girshick, & Farhadi, 2015), SSD (Liu et al., 2016), Mask RCNN (He K., 2017), etc. Currently, Li et al. (Li Liu, 2018) summarizes these state-of-the-art object detection network structures with different time periods and their development process, and Lee (hoseong, 2018) presents most of these model implementation according to the development process.

However, deep neural network-based methods require a large amount of training data and high-performance computing power, which is difficult to obtain in our scenario. The generalization capabilities of these models in our different application scenarios are also uncertain. Additionally, these deep neural network methods can detect multi-object regions, but it is also difficult to further determine the final target region from candidate object regions. Based on the idea of first generating candidate regions and then selecting the target regions, i.e., region proposal idea (Girshick et al., 2014), we propose a selection procedure of the target regions through different motion features of different motion states in our scenario.

**Optical flow.** The latest advances in optical flow are combined with deep neural networks for fast and end-to-end object detection and object tracking (Zhu, Wang, Dai, Yuan, & Wei, 2017; Zhu, Xiong, Dai, Yuan, & Wei, 2016), because optical flow can reflect temporal information of motion features between video frames. However, all of these optical flow methods are based on three major assumptions, which requires that the motion must confirm the following conditions and limits its application scenario: (1) the luminance of the corresponding pixel between adjacent frames is constant; (2) the motion of the corresponding pixel between adjacent frames is slow; and (3) the motion direction of neighboring pixels is similar (Baker & Matthews, 2004). The pyramid method overcomes this obstacle in some extent (BOUGUET, 1999), but it is still impossible to distinguish the temporal and spatial state of the movement from beginning to end.

In contrast, we propose RF-KLT algorithm based on the reference idea, i.e., we do not assume the change between consecutive frames, but measure the change of optical flow between the current frame and a fixed reference frame. Experimental results show that the proposed RF-KLT algorithm can extract robust motion features for motion state detection in a fixed region.

**Time series analysis.** The purpose of time series analysis is generally time series prediction. We classify time series prediction methods into three categories: time series prediction with single target sequence (Xu, Han, Chen, & Qiu, 2018; Zhou et al., 2016), time series prediction combined with external feature sequence of current time (Dipietro, Navab, & Hager, 2017; Qin et al., 2017), and time series prediction combined with external feature sequence of past time (Yunzhe Tao, 2018; Zheng, Liang, Ke, Zhang, & Yi, 2018). Currently, machine learning algorithms, including deep neural networks, are widely used in time series prediction, and it is very important to use optimization algorithms (e.g., particle swarm optimization and genetic algorithm) to optimizing multiple parameters (Sun, Li, Li, Huang, & Li, 2017).

In general, most methods, i.e., the second and third categories, need to be combined with external features that affect the target sequence to achieve more accurate predictions. Actually, time series analysis of single target sequences is more difficult because it does not extract relevant information from other external features. Also, because of the ambiguity of the motion features of the working state, i.e., on or off, we propose a method to construct a feature dataset which is only based on the numerical distribution characteristics of the single target sequence. We directly convert the numerical distribution of the time series extracted from the video into a two-dimensional training dataset. The experimental results show that this method has achieved good results in the feature classification problem of time series.

## 3. Materials and methods

For each specific surveillance camera with different heights and distances in a practical scenario, experiments are required to train a corresponding model. In short, the purpose of object region detection is to detect the aerator region, and the purpose of working state detection is to train the classifier to judge the working state. Notably, the module of work state detection extracts features from the object region in the reference frame. The relationship and purpose of each module are shown in Fig. 1, and the details of each module are given in Section 3.2 and 3.3.

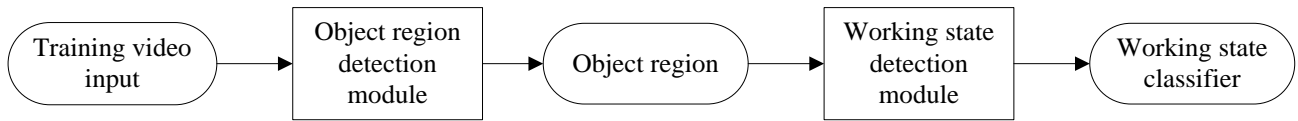

Fig. 1. The relationship and purpose of each module

### 3.1. Video dataset

The video dataset of this article mainly comes from aerator videos which are within 5 minutes and extracted in real time by the Internet of Things (IoT) monitoring platform in Xiongan New Area, China. The presented videos are 8 typical complex scenarios chosen from the video dataset for detecting working state. These videos are from 3 different types of surveillance camera, in which 2 videos are photographed by the camera used to simulate some scenes. The description of video scenes is shown in Table 1. In Table 1, the height of surveillance cameras relative to the object region is divided into three levels: high (H, 8m), medium (M, 5m) and low (L, 2-3m); and the relative distance is also divided into three levels: far (F, >300m), middle (M, 15-70m), and near (N, <15m). It also shows information including time, weather, video parameters, and background description. Notably, the video dataset for detecting object region is the short video clipped in the corresponding video, in which the working state is changed once, and the duration is between 10-20 seconds.

Table 1. Raw dataset description

| Video number | Height | Distance | Frame rate （FPS） | Resolution | Weather | Time | Background description |
|---|---|---|---|---|---|---|---|
| 1 | M | M | 25 | 720*1280 | overcast | a.m. | Ripple, drainage. |
| 2 | M | M | 25 | 720*1280 | sun | p.m. | Direct sunlight, surveillance camera automatically set black and white mode. |
| 3 | M | M | 25 | 720*1280 | overcast | a.m. | Human walking, ripple, drainage. |
| 4 | H | M | 12 | 352*640 | overcast | p.m. | Occlusion. |

| 5 | H | M | 12 | 352*640 | overcast | a.m. | Occlusion, human walking. |
| 6 | L | N | 25 | 1080*1920 | sun | p.m. | Using a camera to simulate a close range scene. |
| 7 | L | N | 25 | 1080*1920 | overcast | a.m. | Using camera to simulate; the strong wind causes camera shake, branch shaking and ripple. |
| 8 | H | F | 25 | 352*640 | overcast | p.m. | Rocking branch, ripple. |

In order to prove the robustness of the model, the pedestrian interference was added to these aerator videos, and the videos were selected under complex scenes such as complicate weathers, different time periods, strong winds, and surveillance camera jitter, etc. Because of the complexity of the working environment, e.g., the illumination changes gradually or suddenly, video datasets are also augmented artificially by adding noise and varying brightness in various proportions. The gray change formula is shown in Formula (3-1).

$$p'(x,y) = \begin{cases} p(x,y) + v & (0 \leq p(x,y) + v \leq 255) \\ 0 & (p(x,y) + v < 0) \\ 255 & (p(x,y) + v > 255) \end{cases} \quad (3-1)$$

Where p(x, y) is the original gray value at the image pixel (x, y), v is the amount of gray change, $p'(x,y)$ is the changed gray value at the image pixel (x, y). In this study, the augmented video dataset was determined by changing the ratio of gray and noise within twice the area of object region. The random number is used in the ratio change. In Table 2, all augmented datasets are based on video 1, and it also shows the random ratio, step size, signal-noise ratio (SNR), and whether the reference frame makes the same change. For example, for video P3, the gray of the reference frame and current frame in the video is increased by 40 in all rows and all columns of the image pixel, and salt-and-pepper noise with the SNR in the range of [0.01, 0.1] is randomly added with 0.01 step size changing in all rows and columns of the image pixel.

Table 2. Augmented dataset description

| Augmented video number | Raw video number | Change ratio of rows(step size is 0.1) | Change ratio of columns(step size is 0.1) | Change range of gray(step size is 1) | SNR of salt and pepper noise(step size is 0.01) | Whether reference frame makes the corresponding change |
|---|---|---|---|---|---|---|
| P1 | 1 | [0:1] | [0:1] | [-80,80] | [0.01,0.1] | Yes |
| P2 | 1 | [0:1] | [0:1] | +40 | [0.01,0.1] | Yes |
| P3 | 1 | All | All | +40 | [0.01,0.1] | Yes |
| P4 | 1 | All | All | +80 | [0.01,0.1] | Yes |

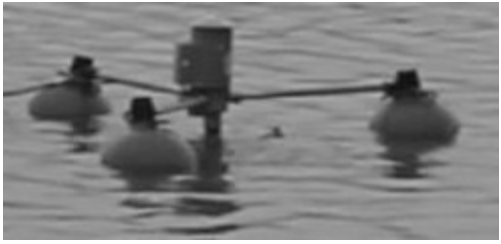 (a)Gray image of object region

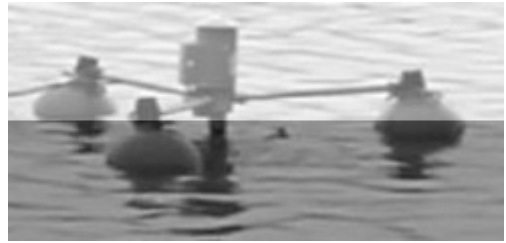 (b)Gray value of the front 1/2 rows increases by 80

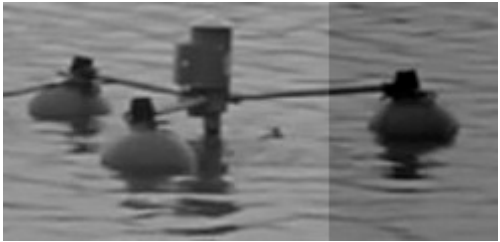 (c)Gray value of the post 1/3 columns reduces by 30

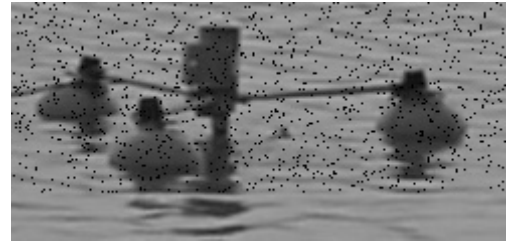 (d)Random salt and pepper noise in the front 4/5 rows(SNR=0.04)

Fig. 2. Example of artificially augmented dataset (the aerator is shown in the figure)

*3.2. Object region detection*

Because the object region determined artificially will lead to the detection of extra feature points and affect detection effect, this paper proposes an automatic detection method for aerator region. The method extracts integrated features from the motion feature and gray feature in inter-frames, and the contour feature in key frame. This method detects the object region in three steps, which is summarized as the detection of maximum contour regions, candidate regions and object region: (1) an adaptive Gaussian mixture model is used to detect the maximum contour region of each frame; (2) the candidate region is detected from the maximum contour regions by integrating the gray feature in inter-frames and the contour feature in key frame; (3) the object region is detected through the candidate regions by calculating the maximum inter-class interval of the motion features dataset based on RF-KLT algorithm. The flow chart of this section is shown in Fig. 3.

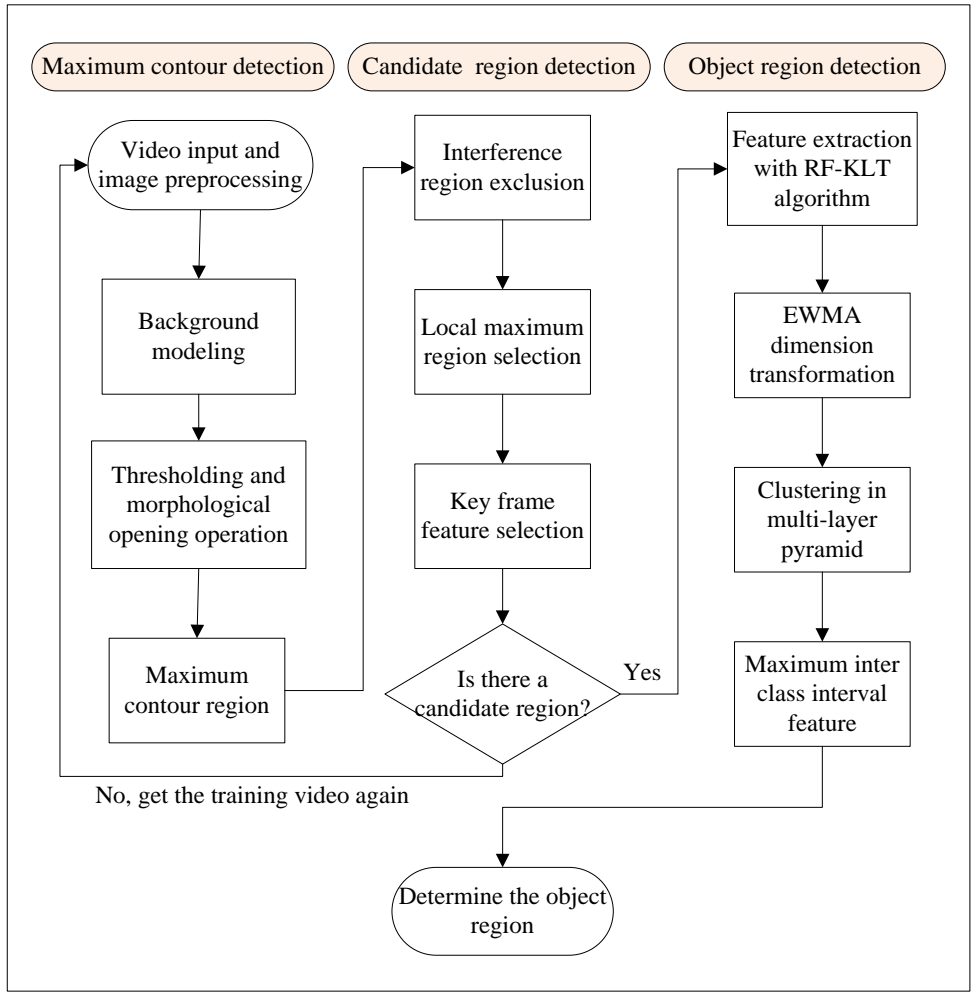

Fig. 3. The algorithm flow of object region detection

3.2.1. Maximum contour detection

This research detects the set of maximum contour regions by an adaptive Gaussian mixture model (Zivkovic, 2004; Zivkovic & Ferdinand, 2006). Background is usually a fixed region with less change, and it is usually assumed that the background can be described by a statistical model. The Gaussian mixture model uses multiple weighted and mixed Gaussian models to analysis background characteristics. Next, the foreground can be detected by marking the part of the image that does not conform to this background model. The Gaussian mixture model (Kaewtrakulpong & Bowden, 2002) can detect the region with small change in the video image, and the adaptive Gaussian mixture model can increase the detection speed and increase the robustness under illumination and ghost regions (Sobral & Vacavant, 2014). The next two steps of the maximum contour region detection are based on the assumption that the aerator region can be detected by an adaptive Gaussian mixture model in the training video.

In this step, Gaussian smoothing is performed on each frame of the video as a preprocessing step firstly. The size of the Gaussian kernel is 5*5. The main purpose of Gaussian smoothing is to avoid detecting some inconspicuous corners. Secondly, an adaptive Gaussian mixture model is used for foreground detection of each frame. It models each pixel with an adaptive number of Gaussian distributions, using the color values of these pixels in the length of the time in the video as a mixed weight, because the color of the background generally lasts the longest and is more static. Thirdly, threshold each frame to obtain a binary image, this research selects a fixed threshold of 240, which can avoid interference in shadowed regions. Fourthly, each frame is subjected to morphological operations to eliminate small regions and separate more obvious foregrounds. The advantage is to smooth the boundary of the larger object region without significantly changing the boundary of these regions. Finally, contour detections are performed on these foregrounds to obtain a region with the largest area from each frame in the video, which constitutes the set of the maximum contour regions.

3.2.2. Candidate region detection

There are too many regions in the set of maximum contour regions, but the areas of the maximum contour regions in most video frames are very small. By excluding these interference regions, the model training speed can be accelerated and the accuracy will be improved. As shown in Fig. 4, the regularity curve between the areas of maximum contour regions and the number of frames obeys the law shown in Formula (3-2). In maximum contour regions, the first 1/4 regions with the largest area are selected to exclude interference regions with a small area. In order to avoid interference and repeated detection in adjacent regions, regions where the centroid distance of the contour is larger than a threshold are selected from the rest of maximum contour regions. The local maximum region is selected as a candidate region. 1/10 of minimum value from video frame resolution (e.g., 720/10) is used as the threshold for excluding adjacent regions.

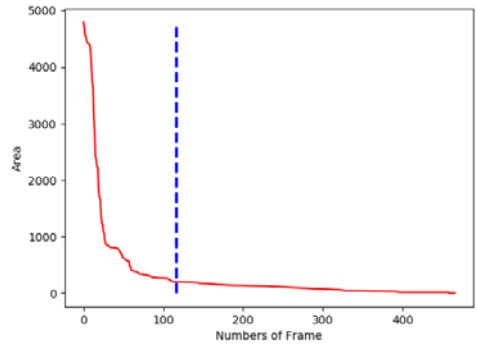

Fig. 4. Example of area distribution of maximum contour regions

In short, the object region belongs to two states which include nonworking state and working state. There are a lot of feature points in both states, because the body of aerators and the edge of the spray are both changed strongly on the gray value. The interference region, e.g., branch, is also a region of strongly gray change. In addition, the features of single frame image can not detect the motion features of object region. Hence, the frame that appears in the maximum contour region is selected as a key frame for features extracted in this region. Also, the first frame of the video when the aerator is closed is defined as the reference frame. Compared with maximum contour region in the key frame and corresponding region in the reference frame, features of the gray change in the spray region are extracted. In order to avoid the influence of illumination and improve the detection accuracy of small region, the feature of contour area is considered comprehensively. The feature function of the key frame and the reference frame is determined finally in Formula (3-3).

$$y = \frac{k}{x} (k > 0) \qquad (3-2)$$

$$F = \frac{\sum_0^x \sum_0^y (I(x,y) - J(x,y))}{A_{max}} \qquad (3-3)$$

Where F is the feature value extracted in the selected region, $I(x,y)$ is the gray value of image point $(x,y)$ in the selected region of key frame, $J(x, y)$ is the gray value of image point $(x, y)$ in the selected region of reference frame, and $A_{max}$ is the maximum contour area in the selected region of key frame. F feature can be used to detect the spray region caused by the work of aerators, and it is also robust to the monitoring of small object region. If the candidate region is smaller, the area of detected contour is smaller and the corresponding F increases. This paper finally selects the set of regions where the F feature value is larger than the average value in the maximum contour regions as candidate regions.

3.2.3. Object region detection

The RF-KLT-based feature extraction method and the maximum inter-class interval measurement method are both used to determine the final object region from candidate regions. In KLT tracker (Shi, 1994) and its pyramidal implementation (BOUGUET, 1999), consider an image point $\mathbf{u} = [x \ y]^T$ on the frame $I$. The objective of KLT tracker is to find the location $\mathbf{v} = \mathbf{u} + \mathbf{d} = [x + d_x \ y + d_y]^T$ on the next frame $J$ such as the gray value $I(\mathbf{u})$ and $J(\mathbf{u})$ are "similar". The vector $\mathbf{d} = [d_x \ d_y]^T$ is the optical flow at point $\mathbf{u}$. It is essential to define the notion of similarity in a two-dimensional neighborhood sense because of the aperture problem. The optical flow $\mathbf{d}$ is defined as the vector that minimizes the residual function $\epsilon$ that defined in Formula (3-4), where the similarity function is measured on an image neighborhood of size $(2w_x + 1) \times (2w_y + 1)$. The value of $w_x$ and $w_y$ are integers, for which the typical values are 2, 3, 4, 5, 6, 7 pixels.

$$\epsilon(\mathbf{d}) = \epsilon(d_x, d_y) = \sum_{x=u_x-w_x}^{u_x+w_x} \sum_{y=u_y-w_y}^{u_y+w_y} (I(x,y) - J(x+d_x, y+d_y))^2 \qquad (3-4)$$

The objective of pyramidal implementation of KLT tracker is also to find the location $\mathbf{v} = \mathbf{u} + \mathbf{d} = [x + d_x \ y + d_y]^T$ on the next frame $J$ such as the gray value $I(\mathbf{u})$ and $J(\mathbf{u})$ are "similar", but the point $\mathbf{v}$ is not in the image neighborhood of size $(2w_x + 1) \times (2w_y + 1)$, and is far away from the point $\mathbf{u}$ in a larger range. It is preferable to have $d_x \leq w_x$ and $d_y \leq w_y$ in Formula (3-4) to find the accurate and robust point $\mathbf{v}$, so the pyramid is used instead of expanding the size of image neighborhood. The optical flow $\mathbf{d}^L = [d_x^L \ d_y^L]^T$ at level L in pyramid is defined as the vector that minimizes the new residual function $\epsilon^L$ that defined in Formula (3-5).

$$\epsilon^L(\mathbf{d}^L) = \epsilon^L(d_x^L, d_y^L) = \sum_{x=u_x^L-w_x}^{u_x^L+w_x} \sum_{y=u_y^L-w_y}^{u_y^L+w_y} (I^L(x,y) - J^L(x + g_x^L + d_x^L, y + g_y^L + d_y^L))^2 \qquad (3-5)$$

Where $\mathbf{g}^L = [g_x^L \ g_y^L]^T$ is available from the computations done from level L to level L + 1 in pyramid, which is computed by Formula (3-6).

$$\mathbf{g}^{L-1} = 2(\mathbf{g}^L + \mathbf{d}^L) \qquad (3-6)$$

For level L in pyramid, $\mathbf{d}^L$ is computed through the same procedure by Formula (3-5), which searches the finest point $\mathbf{v}$ to minimizes the functional $\epsilon^L(\mathbf{d}^L)$ in the neighborhood with constant size $(2w_x + 1) \times (2w_y + 1)$. This procedure goes on until the bottom level is reached (L = 0), and the initial value of the top level (L = $L_{max}$) in pyramid is initialized to zero.

$$\mathbf{g}^{L_{max}} = [0 \ 0]^T \qquad (3-7)$$

Therefore, the final optical flow $\mathbf{d}$ at point $\mathbf{u}$ is then defined in Formula (3-8).

$$\mathbf{d} = \mathbf{g}^0 + \mathbf{d}^0 = \sum_{L=0}^{L_{max}} 2^L \mathbf{d}^L \qquad (3-8)$$

The objective of pyramidal implementation of RF-KLT tracker is the same as the above two methods, but it is changed to find more obvious motion feature, and guarantees the accuracy and robustness of feature point tracking. For each candidate region, RF-KLT algorithm proposed in this paper is divided into three steps, the schematic diagram is shown in Fig. 5. (1) Shi-Tomas corner detection method is used to find the corners (point **u**) in the fixed reference frame. The Shi-Tomas algorithm (Lucas & Kanade, 1981) determines the strong corner point through finding the maximum value from the minimum eigenvalue of each two eigenvalues in the pixel gradient matrix. It can ensure that the number of detected feature points is within a reasonable range, and the feature points in the object region all fall on the body of aerators. (2) Lukas-Kanade algorithm is used to find the matching corners (point **v**) in current frames corresponded reference frame. The search range of the matching corner points starts from the center of object region, and the radius of the diagonal of the object region is selected as radius. Lukas-Kanade algorithm instructs how this method is used for feature point tracking and has evolved into a practical KLT tracker (Shi, 1994)

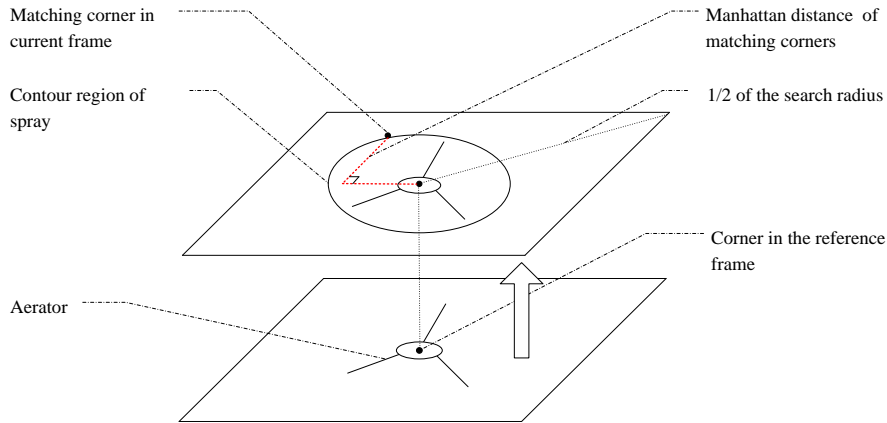

Fig. 5. Dist feature extraction based on RF-KLT algorithm

(3) The average distance between matching corners of each frame is extracted as the final feature. Manhattan distance is used to measure this distance between matching corner points. The extraction of Dist features in each frame based on RF-KLT algorithm is shown in Formula (3-11).

$$\epsilon^L(\mathbf{d}^L) = \epsilon^L(d_x^L, d_y^L) = \sum_{x=u_x^L-w_x}^{u_x^L+w_x} \sum_{y=u_y^L-w_y}^{u_y^L+w_y} (I_0^L(x,y) - J^L(x + g_x^L + d_x^L, y + g_y^L + d_y^L))^2 \quad (3-9)$$

$$\mathbf{d} = \mathbf{g}^0 + \mathbf{d}^0 = \sum_{L=0}^{L_{max}} 2^L \mathbf{d}^L \quad (|\mathbf{d}| \leq 2\sqrt{(l/2)^2 + (h/2)^2}) \quad (3-10)$$

Compared with the conventional KLT algorithm, the RF-KLT algorithm is improved in three aspects. The calculation procedure is shown in Formula (3-9) and Formula (3-10), and schematic diagram and result formula as shown in Fig. 5 and Formula (3-11) respectively. Where $I_0^L(x,y)$ is a point $\mathbf{u} = (x,y)$ in fixed reference frame in the level L of pyramid, and $l$ and $h$ are the length and width of the candidate region respectively. The main procedure includes: (1) Finding the best matching point of the corresponding corner point in the fixed reference frame, which means that $I_0^L$ is not changed along with $J^L$ changed. The goal of this improvement is to obtain the movement state change characteristics with more distinct discrimination, and this is the most critical improvement. (2) Finding the best search radius under the multi-level image pyramid. It means that the range (**d**) of searching matching point by optical flow method is limited. The goal of this improvement is to ensure that the matching corner is not lost and robust. (3) Using Manhattan distance to measure the distance between the corner points (point **u** and point **v**), which is also to obtain more obvious movement characteristics. Notably, the precondition for matching corners using optical flow method based on the fixed reference frame is as follows: a multi-level pyramid and a fixed region. These two conditions ensure that the three basic assumptions of the optical flow method are still confirmed and the first and second assumptions are expanded.

$$\text{Dist} = \begin{cases} \left(\frac{1}{i}\right) * \sum_{0}^{i} |x_{i1} - x_{i0}| + |y_{i1} - y_{i0}| & (0 < i \leq 5) \\ 0 & (i = 0) \end{cases} \quad (3-11)$$

Where Dist is the distance feature value between matching corner points, $(x_{i0}, y_{i0})$ is the coordinates of the i-th corner point in the reference frame, and $(x_{i1}, y_{i1})$ is the coordinates of the i-th matching corner point in the current frame corresponds to the reference frame, and i is the number of corners. In order to ensure the real-time detection, and according to the number of strong corners on the body of aerators, the most obvious 5 corners with maximum eigenvalue are selected. Also, $w_x$ and $w_y$ are both 7 pixels in this study.

Dist feature is an unlabeled dataset that changes with the number of frames, and it can be considered as time series data. In order to smooth the data and perform dimensional transformation, the exponentially weighted moving average (EWMA) model (Kim & Chang, 2018) is used to process original dataset, and its calculation formula is shown in Formula (3-12). EWMA model maintains coherence of the inter-frame feature value in a fixed length window. Furthermore, the most important is that the two-dimensional feature reconstructed by deleting the time dimension is more robust to distinguish. In the constructed dataset, one class of data is close to the origin point, and one class of data is far from the origin point. The two classes of datasets constructed in this way also have good flatness in shape.

$$y_t = \frac{x_t + (1-\alpha)x_{t-1} + (1-\alpha)^2 x_{t-2} + \cdots + (1-\alpha)^t x_0}{1 + (1-\alpha) + (1-\alpha)^2 + \cdots + (1-\alpha)^t} \quad (3-12)$$

Where $y_t$ is the EWMA feature corresponding to the Dist feature in the current frame; $x_t$ is the Dist feature in the current frame, which is the Dist feature in the last frame of the sliding window. $x_{t-1}$ is the Dist feature in the previous frame of the sliding window; etc. $x_0$ is the Dist feature in the first frame of the sliding window; $0 < \alpha < 1$, and $\alpha = \frac{2}{s+1}$, where $s \geq 1$, and s is the frame rate of video. t is the length of the sliding window, and its selection is also automatically obtained based on the frame rate of video.

After data smoothing and dimensional transformation, the dataset has a distinct distribution rule of two classifications. Because the dataset is flat and the numbers of class are known, the K-means clustering method (K=2) can accurately obtain the centroid of each class of dataset and thus calculate the inter-class interval feature (Arthur & Vassilvitskii, 2007). Details of K-means clustering method are given in 3.3.2 section. The maximum inter-class interval feature under the multi-level pyramid can prevent the size of object region from affecting experimental results. As shown in the Formula (3-13), the region with a largest centFeature is selected as the final object region from the candidate regions. The centFeature under the high level pyramid can avoid a small feature value between two classes in the small region. At the same time, by reducing the proportion of distance features under the high level pyramid, large feature value between two classes in the large region can be avoided.

$$\text{CentFeature} = \sum_{0}^{i} \frac{Diff}{2^i} \quad (0 \leq i \leq 4) \quad (3-13)$$

Where CentFeature is inter-class interval feature of the selected candidate region. Diff is the difference of abscissa values between two class centroids under the i-th pyramid of the selected region, and i is the number of pyramid levels, increasing from 0 to 4 levels.

*3.3. Working state detection*

The purpose of this section is to achieve automatic detection of aerator working state. The algorithm flow of working state detection is shown in Fig. 6. In this module, the main procedures include feature extraction, feature dataset construction (including EWMA data processing and dimension transformation), data labelling, and classifier training. The details of RF-KLT algorithm and EWMA model are described in 3.2.3 section. In the reference frame of RF-KLT algorithm, corner detection is performed in the object region obtained by object region detection module.

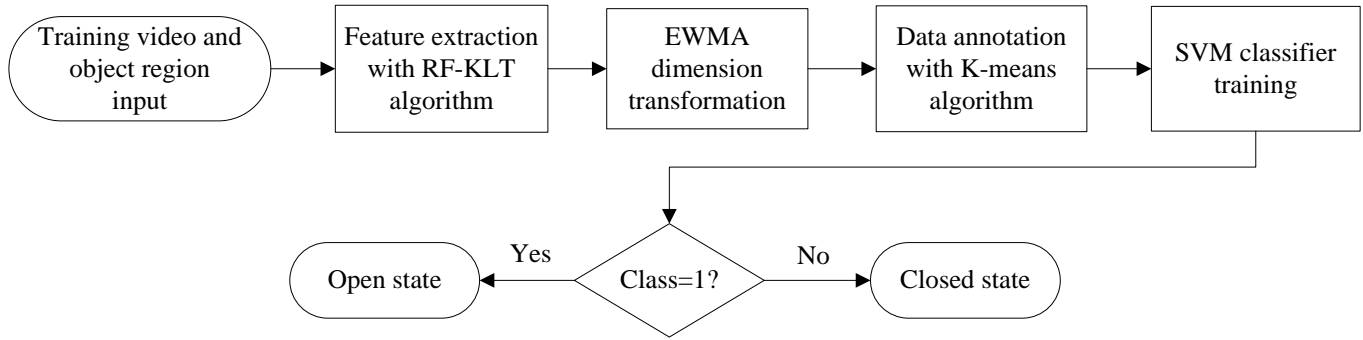

Fig. 6. The algorithm flow of working state detection

3.3.1. Feature extraction and dataset construction

The extraction of image features in computer vision is a very critical step. The purpose of RF-KLT algorithm is to extract robust features for the motion state change. Note that corner matching is done under the 4-level pyramid. The high level pyramid can increase the search range of corner matching, and get larger features of corner distance variation, which increase the discrimination of datasets with two classes. The conventional optical flow method is based on the detection of adjacent frames, but the corner change features are not obvious in the motion state detection of fixed region. Therefore, the proposed RF-KLT algorithm can be used to extract more obvious characteristics of motion state change in scenes of brightness change tempestuously and large scale motions. This algorithm can also be used in other conditions to detect motion state change in a fixed region.

The EWMA model is used to transform time series data into the same dimensionality to construct a dataset that is smoother and more distinguishable, which can avoid the influence of single error on the experimental results, and can also reduce the influence of data points from working state change process on the experiment. The constructed dataset has flat and highly discriminating characteristics.

3.3.2. Unlabeled data annotation

Clustering algorithm is widely used in the learning of unlabeled samples, which is used to reveal the inherent laws of data and provide a basis for further data analysis. Clustering algorithm is a typical unsupervised learning algorithm, which is mainly used to automatically cluster similar samples into one class. The number of classes and inter-class characteristics of the datasets constructed in this paper are known, including the on and off of two classes of feature data. One class dataset is closer to the origin point, and another is far from the origin point. In order to achieve data labeling, and because of the flatness and aggregation shape of dataset distribution, K-means clustering methods are used to annotate data (Arthur & Vassilvitskii, 2007).

The idea of the K-means algorithm is simple. Firstly, the constant K is determined, which means the number of clusters. In this study, K=2. Then the starting point is selected randomly as the centroid. Secondly, measuring the similarity between each sample and the centroid by Euclidean distance, and placing the sample points in the most similar class. Thirdly, recalculating the centroid of each class and repeating the procedure until the centroid does not change or reach the upper limit of iterations. Finally, the class to which each sample belongs and the centroid of each class are determined.

In this study, label 1 is assigned to the dataset that is far from the origin point corresponding to the open state, and label 0 is assigned to the dataset that is close to the origin point corresponding to the closed state. The feature data in the state change process and the feature data due to the corner matching errors are determined based on the distance from the centroids of the two classes. This enables automatic labeling of unlabeled datasets. In order to avoid classifying high-error data into one class, the feature dataset constructed by the two working states of aerators must be a balanced dataset in training videos. This study sets the ratio of two classes of dataset between 1/5-5. The object region is very small, the change of features is also small. Hence, if the corner matching is not stable, the mismatched feature value will be too large. Then the error data becomes a class of data, and the trained classifier cannot correctly detect the open state. Therefore, the balance of training dataset can avoid such errors.

3.3.3. Classifier training

Through the feature extraction from different working states, the construction of datasets, and the labeling of unlabeled data, the final dataset is a triple, which includes two-dimensional feature data and their corresponding labels. The SVM algorithm has a complete theoretical basis and good classification ability, and is widely used in the data classification with limited samples (Wu, Lin, & Weng, 2004). This research constructs a dataset that is linearly separable, so the linear SVM algorithm is used to effectively classify the dataset.

At last, the application procedure of our methods is divided into three steps: (1) the corner distance feature (Dist feature) of real-time frame is extracted according to the matching corner points in the object region of the reference frame; (2) constructing two-dimensional data using the EWMA model; and (3) classifying data according to trained classifiers. If the class label is one, it indicates that the aerator is currently in the working state; and if the class label is zero, the aerator currently is in the closed state.

**4. Results and discussions**

*4.1. Software and algorithm parameter*

In this study, the software is programmed in Python 3.6.3 programming language using the Numpy and Pandas scientific computing libraries. Opencv 3.3.1 in python is used for the basic image processing and video analysis (OpenCV, 2017). Scikit-learn 0.19.1 in python is used for machine learning algorithms comparing (Scikit-learn, 2017). The algorithms development and experiments are based on the 64-bit Intel 3$^{rd}$-Generation Core i5 CPU (i5-3230M, 2.60GHz). This is a personal laptop with a common configuration, indicating that the experimental results have a good reference value in the ordinary monitoring system configuration.

The precision, recall and f1-score are used to evaluate the results of working state detection, and they are defined as Formula (4-1), Formula (4-2) and Formula (4-3).

$$\text{precision} = \frac{tp}{tp+fp} \qquad (4-1)$$

$$\text{recall} = \frac{tp}{tp+fn} \qquad (4-2)$$

$$f1 = 2 * \text{presion} * \text{recall}/(\text{precision} + \text{recall}) \qquad (4-3)$$

Where $tp$ is the number of correct samples to be classified correctly, $fp$ is the number of incorrect samples to be classified correctly, and $fn$ is the number of incorrect samples to be classified incorrectly. In the comparison experiment of the algorithm, the key parameters and principled literature in the involved algorithms are given in Table 3. Most of these algorithms are from OpenCV library and Scikit-learn library respectively. The specific meaning and details of the parameters in the algorithm refer to the official documentation, and abbreviations of some algorithm are used in this study.

Table 3. Algorithms and key parameters in the comparison of experiments

| Algorithm | Abbreviation | Key parameters in this study | Reference | Document |
|---|---|---|---|---|
| Support vector machine | SVM | C=1, kernel: linear | (Wu et al., 2004) | Scikit-learn, (Scikit-learn, 2017) |
| Linear regression | LR | C=1, solver: liblinear | (Schmidt, Roux, & Bach, 2013), | |
| Linear discriminant analysis | LDA | solver: svd | (Fan, Lei, & Li, 2008) | |
| K-Nearest Neighbor | KNN | n_neighbors=5, weights: uniform | (Munther, Razif, Abualhaj, Anbar, & Nizam, 2016) | |
| Classification and regression trees | CART | criterion: gini | B. LI, Friedman, Olshen, & Stone, 1984) | |
| Naive bayes | NB | priors: None | (He, Zhang, Li, & Wang, 2011) | |
| K-means clustering | / | n_clusters=2, init: k-means++, max_iter=300 | (Arthur & Vassilvitskii, 2007) | |
| Spectral clustering | / | n_clusters=2, affinity: rbf, gamma=1.0, assign_labels: kmeans | (Luxburg, 2007) | |
| Agglomerative clustering | / | n_clusters=2, linkage: ward | (Szekely & Rizzo, 2005) | |
| Adjacent frame difference | AFD | Gaussian smoothness(kernel: 5*5), threshold of binary image:25 | / | / |
| White region selection in HSV space | WRS-HSV | Gaussian smoothness(kernel: 5*5), H: 0-180, S: 0-30, V: 220-255 | / | |
| Fixed background subtraction | FBS | Gaussian smoothness(kernel: 5*5), threshold of binary image: 25 | / | |
| Gaussian mixture background model | GMB | Gaussian smoothness(kernel: 5*5), threshold of binary image: 240 | (Kaewtrakulpong & Bowden, 2002) | |
| K-Nearest Neighbor algorithm | KNN-B | history=30, Gaussian smoothness(kernel: 5*5), threshold of binary image: 240 | (Zivkovic & Ferdinand, 2006) | Opencv, (OpenCV, 2017) |
| Adaptive Gaussian mixture background model | AGMB | Gaussian smoothness(kernel: 5*5), threshold of binary image: 240 | (Zivkovic, 2004; Zivkovic & Ferdinand, 2006) | |

*4.2. Comparison and discussion of experiments*

4.2.1. Comparison between RF-KLT algorithm and KLT algorithm

Compared with the classical KLT algorithm, the RF-KLT algorithm proposed in this paper has a significant discrimination degree of motion feature in the fixed object region. The comparison of the differences in the motion state change between KLT and RF-KLT algorithm is shown in Fig. 7. For the video with same work state change, the feature change relative to the reference frame is more obvious than that of the adjacent frame. KLT algorithm has continuous corner motion, and its feature change between matching corner points is small and the distinction degree is not high. However, the corners in the reference frame are fixed in RF-KLT algorithm. Specifically, when the working state is unchanged, the matching corner point is itself; and when the working state changes, there is no matching corner point in the spray region. Then, the matching corners will follow the boundary between the spray and the calm water surface. When the spray region is stable in the maximum region, the matching corner position is fixed. Therefore, Dist feature based on the RF-KLT algorithm have a good discrimination degree on the change of motion state. At the same time, the RF-KLT algorithm selects the Shi-Tomas corner with the largest eigenvalue, and it has a fast speed of 0.002 FPS in video 1 compared to the Harris corner point (Harris, 1988).

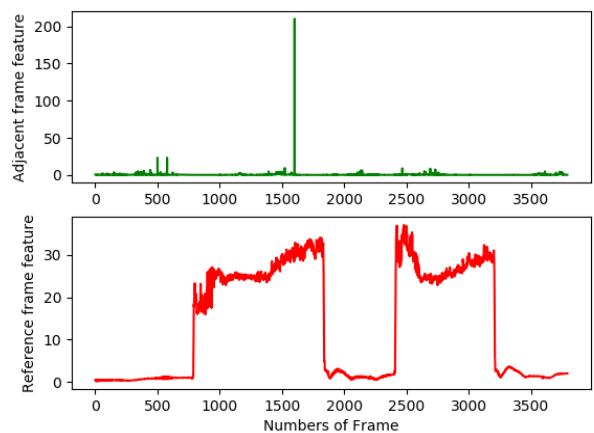

Fig. 7. Comparison of Dist value between conventional KLT algorithm and RF-KLT algorithm (video 1)

Image Pyramid can match feature points in a larger pixel range, and RF-KLT algorithm is also built on the multi-level pyramid. The relationship between the number of pyramid levels and the distance feature of matching corner points is shown in Fig. 8. In Fig. 8, when the pyramid is higher than four levels, the distinguishability of the characteristics no longer increases. Therefore, the RF-KLT algorithm in the module of working state detection is to match the corner points under the 4-level pyramid. The detection time of each frame and the degree of instability of corner matching increase with the increase of the number of pyramid levels, so higher-level pyramid do not need to be constructed without improving feature discrimination.

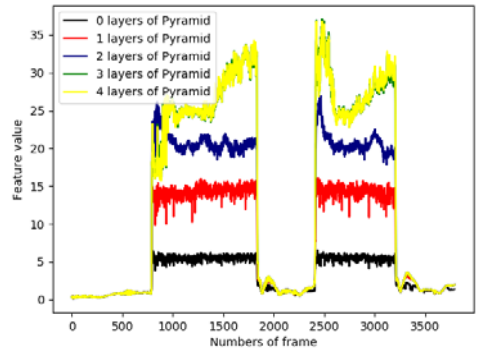

Fig. 8. Comparison of Dist value in different pyramid levels (video 1)

The selection of the reference frame leads to an obvious distinction of the extracted features, so the determination of the reference frame is very critical. The reference frame of the RF-KLT algorithm can be determined in one frame, i.e., any frame of the non-working state of aerators can be selected. Fig. 9 shows the accuracy of working state detection with different reference frame in video 1. The experimental results show that there is no effect on the experimental results to select $1^{st}$, $10^{th}$, $20^{th}$, $40^{th}$ or $80^{th}$ frame as the reference frame, which proves that the RF-KLT algorithm is stable and robust.

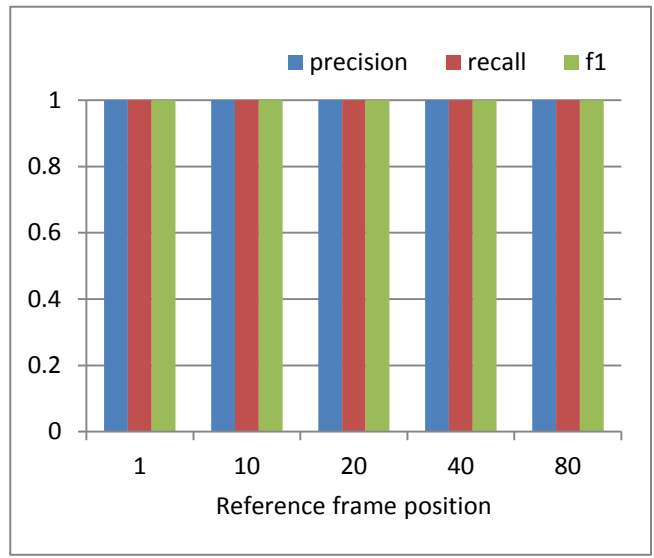

Fig. 9. Comparison of the influence of different reference frame in working state detection module

4.2.2. Comparison with different object detection algorithms

The object region detection of videos in this study is divided into three steps. Table 4 shows the influence of the different object detection method on the final detection result in different video data. In this experiment, the foreground detection algorithm is used in the maximum contour region detection module, and the subsequent two-step detection methods are not changed. This paper compares the influence of AFD, WRS-HSV, FBS, GMB, KNN-B and AGMB (details in Table 3) on the experimental results in the maximum contour regions detection, and the last method is our selected algorithm. In Table 4, the object region is correctly identified in various complex scenarios based on the KNN algorithm and our model, which shows that the principle of regional selection procedure constructed in this paper is completely correct. For simple background, most algorithms can accurately detect the object region. However, in complex background, e.g., the object region is too small or the interference is heavy, the KNN algorithm and our algorithm have good stability. In the time performance of each frame, as shown in Fig. 10, our method is faster than KNN in all scenarios.

Table 4. Comparison of different foreground detection algorithms on object region detection

| Video number | AFD | WRS-HSV | FBS | GMB | KNN-B | AGMB(Ours) |
|---|---|---|---|---|---|---|
| 1 | ☑ |   | ☑ | ☑ | ☑ | ☑ |
| 2 | ☑ | ☑ | ☑ | ☑ | ☑ | ☑ |
| 3 | ☑ | ☑ |   |   | ☑ | ☑ |
| 4 |   | ☑ | ☑ | ☑ | ☑ | ☑ |
| 5 | ☑ |   | ☑ | ☑ | ☑ | ☑ |
| 6 | ☑ | ☑ | ☑ | ☑ | ☑ | ☑ |
| 7 | ☑ | ☉ | ☑ | ☑ | ☑ | ☑ |
| 8 | ☉ | ☉ | ☉ | ☉ | ☑ | ☑ |

(☑ indicates that the final result of object region detection is accurate; ☉ indicates that the final result of object region detection is inaccurate, but the object region is detected in the candidate regions; and empty table indicates that object region is not detected in the maximum contour regions.)

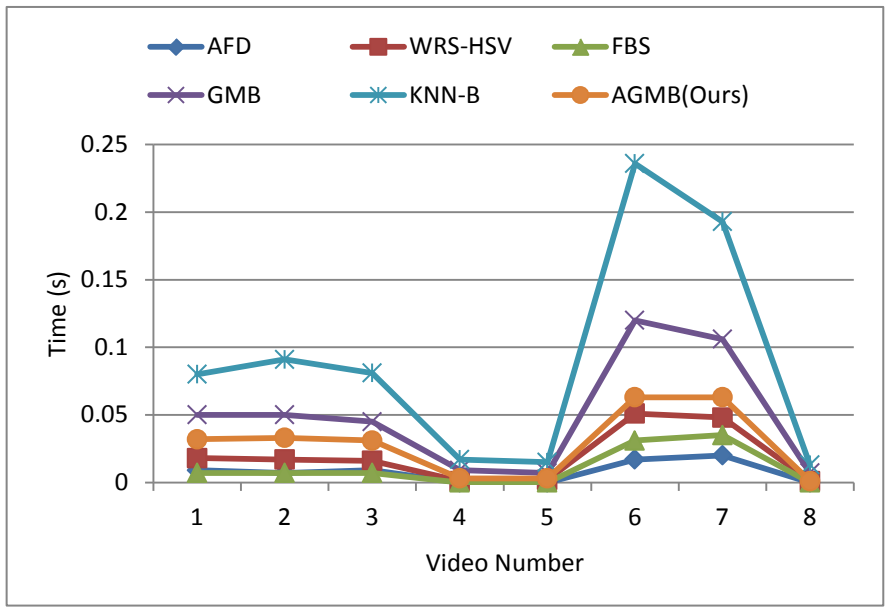

Fig. 10. Speed comparison of different foreground detection algorithms

4.2.3. Comparison with different machine learning algorithms

After the data is labelled by the clustering algorithm, a classifier with high precision and high generalization ability can be trained for detecting the working state. The result of the combination of several clustering algorithms and classification algorithms in video 1 and video 8, which represents simple and complex scenes respectively, is shown in Fig. 11. In this paper, 4/5 of dataset is used as training dataset, and the rest of dataset is used as test dataset. The average precision and standard deviation of the training dataset classification result are determined using a five-fold cross validation method. In Fig. 11, K-means clustering, spectral clustering and agglomerative clustering with ward linkage are used to label data. LR, LDA, KNN, CART, NB and SVM with linear kernel (details in Table 3) are used to build the classifier for the annotated dataset.

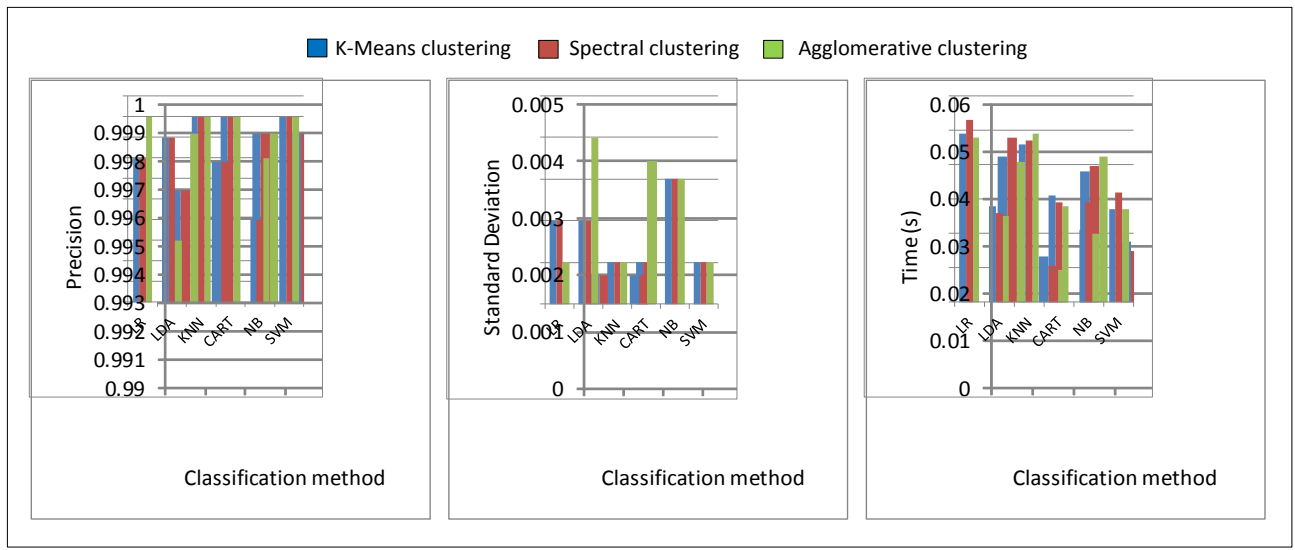

(a) Comparison of different machine learning algorithm combinations in video 1

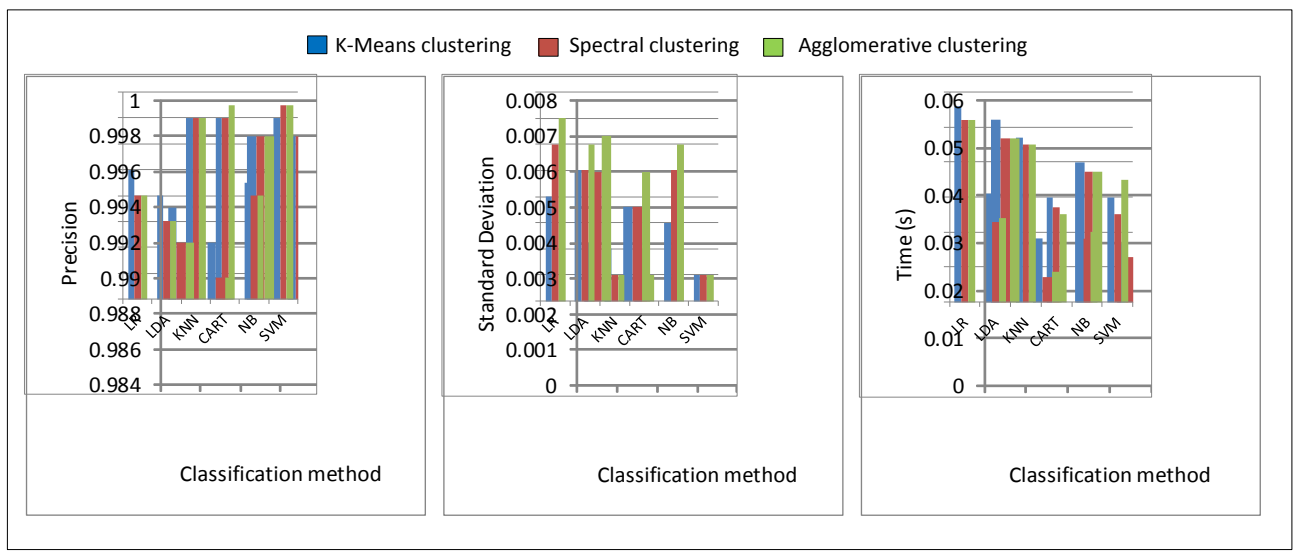

(b) Comparison of different machine learning algorithm combinations in video 8

Fig. 11. Comparison of different machine learning algorithm combinations in typical video dataset

The principle of selecting a clustering algorithm is based on the regularity of dataset: (1) two classes, (2) linear separation and obvious boundaries between classes, and (3) determination of the class labels corresponding to the different working state of aerators. Other clustering algorithms such as density clustering algorithms label data into multiple classes because the number of classes cannot be preset. This study compares three linear classifiers: LR, LDA, and SVM with linear kernel. At the same time, three nonlinear classification algorithms have been added for comparison. From the clustering result of video 1 in Fig. 11(a), there are no obvious differences in the three methods of the result in the clustering algorithm; and in the classification algorithm, the accuracy and standard deviation of KNN, CART, and SVM algorithms are all good. However, the KNN algorithm is worse in time cost, and the CART algorithm is also slightly slower than the SVM algorithm. As shown in Fig. 11(b), for the complex dataset such as the small object region in video 8, the SVM algorithm is also an optimal algorithm in accuracy, standard deviation, and time cost.

The average time performance of each frame used in clustering algorithms is shown in Fig. 12. The time performance of the K-means algorithm is one or two orders of magnitude faster than the other two clustering algorithms. According to the accuracy and time analysis, this paper selects the K-means algorithm for data annotation, and selects the linear SVM algorithm to train the classifier. Notably, the results of algorithm comparison also show the robustness and stability of the feature extraction method and dataset reconstruction method in this study, because the accuracy of dataset classification is closed to 100% in the combination of all the different machine learning algorithms. In effect, the accuracy of our algorithm combination in training dataset is 99.9%.

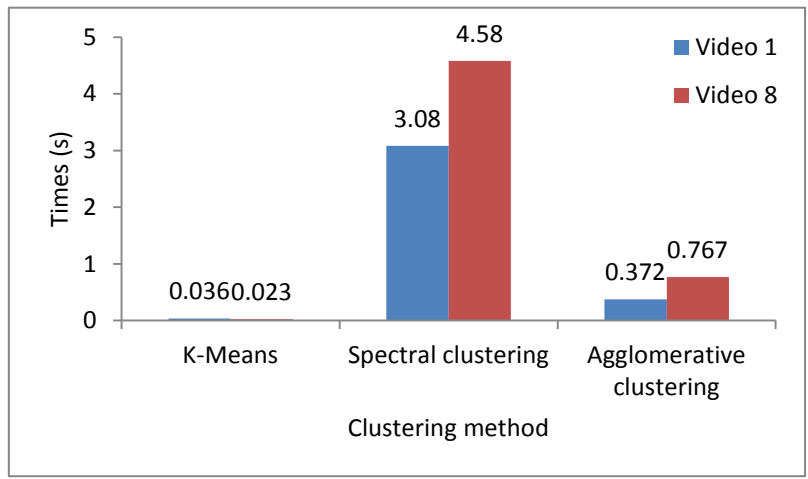

Fig. 12. Speed comparison of different clustering algorithms

As shown in Formula (4-4), in order to avoid labeling the error data as one class, and to prove that the two classes of data labeled by clustering algorithms are closed state and open state respectively, this paper designs a kind of evaluation index—class coefficient (CR), according to the Jaccard coefficient, to measure whether the data is correctly labeled.

$$\text{CR} = \frac{|N_{negative} - N_{openframe}|}{N_{frames}} \quad (4-4)$$

Where $N_{negative}$ is the number of negative samples, i.e., the number of samples near the zero point; $N_{openframe}$ is the frame number when the object region is detected in the training video; $N_{frames}$ is the total number of frames in training video. Note that the CR coefficient is calculated based on working state changed once. It can measure the proportion of a correct sample in the total sample. Specifically, when it is detected that the frame of the object region appears, i.e., the aerator has been opened steadily, the sample data is converted to positive class, thereby the value of $N_{negative}$ is determined. The degree to which the CR is close to zero indicates how accurately the construction dataset was labeled. However, CR cannot be 0, because $N_{openframe}$ is determined when the object region is detected, the matching corner distance feature has reached its maximum value. Accordingly, some of the samples in the state change process have been classified as positive class. Table 5 is the CR value of video 1 and video 8 corresponding to the above comparison experiment of machine learning algorithms, in which the video 8 only intercepts segments whose working state of object region changes once. The result shows that the clustering methods of this paper have higher accuracy in labeling samples, which also shows the obvious distance between classes of the constructed dataset.

Table 5. CR coefficients of different clustering algorithms in typical video dataset

| video number | CR | | |
| --- | --- | --- | --- |
| | K-means | Spectral clustering | Agglomerative clustering |
| 1 | 0.023 | 0.023 | 0.023 |
| 8 | 0.059 | 0.059 | 0.059 |

*4.3. The performance of our expert system*

4.3.1. Object region detection

In Fig. 13, the time of each step in the object region detection module and the accuracy of our algorithm in the test dataset are shown. Time1 is the detection time of each frame in maximum contour region detection step, time2 is the detection time of each frame in candidate regions detection step, and time3 is the detection time of each region in object region detection step. The detection of the maximum contour region takes less time, and the time for selecting the candidate object region is even more negligible. This is because the value of gray change in the F feature of object region is calculated along with the contour area, so this step is only judging result. The proportion of the time taken to obtain the largest inter-class interval under the multi-level pyramid is large, which is why the authors suggest that the training video of the object region detection module should be between 10-20 seconds. Also, another reason is, as the number of training video frames increases, the number of candidate regions increases accordingly, resulting in an increase in the time for determining the final object region from the candidate regions.

The object region detection is mainly to prepare for the detection of work state. The most crucial step in the feature extraction of work state detection is the matching of corner points. The best result is that all detection corners fall on the body of aerators. In this study, the detection performance is evaluated according to Formula (4-1). However, $tp$ is the corner point in object region that is correctly detected and $fp$ is the corner point in object region that was detected incorrectly. As shown in Fig. 11, the object region is fully detectable according to the precision result.

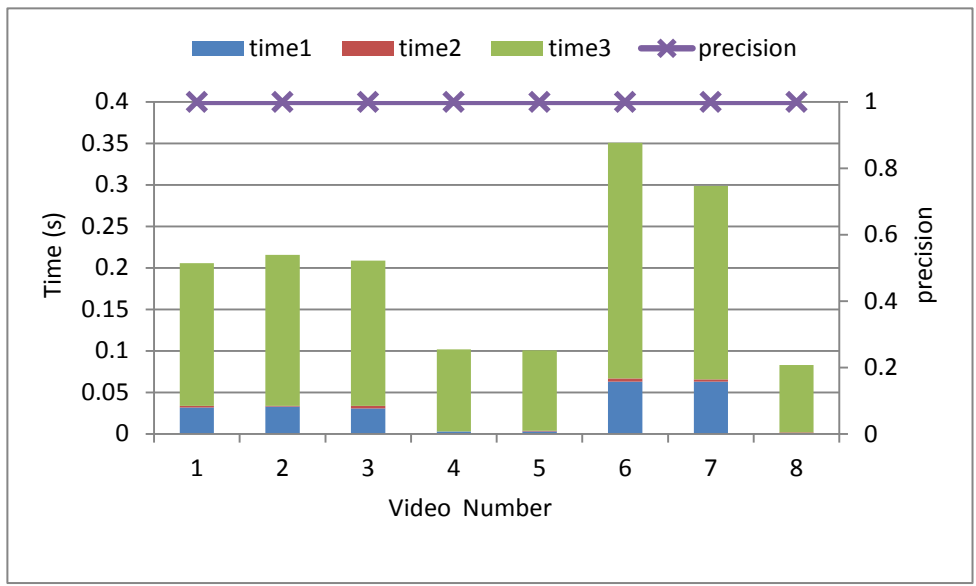

Fig. 13. Time and accuracy of object region detection method for different video datasets

4.3.2. Working state detection

The evaluation indicators, including precision, recall and f1-score, of the working state detection in the video datasets are shown in Fig. 14, and the detection time of each frame is shown in Fig. 15. In Fig. 14, the discriminant model can reach 100% accuracy in test dataset in all scenes that include videos with artificially increasing brightness and noise. This results mean that the features extracted based on the RF-KLT algorithm, the method of datasets constructed, and labeling method are all robust. In Fig. 15, videos 6 and 7 are camera supplement scenes. Due to its highest resolution, the detection time of each frame is longer. In short, the detection time from different types of surveillance cameras is different and the detection time increases with the higher resolution of the surveillance camera. The results show that the detection speed of real surveillance cameras is between 77-333 FPS, which is much higher than 25 FPS of general surveillance cameras, which shows this method is real-time in many complex scenarios. Also, the results of the augmented video dataset artificially prove the stability of our method.

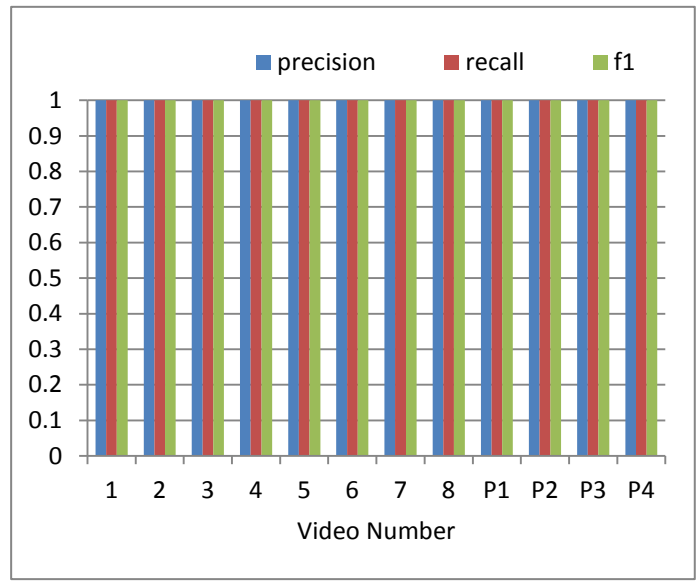

Fig. 14. Performance of working state detection for different video datasets

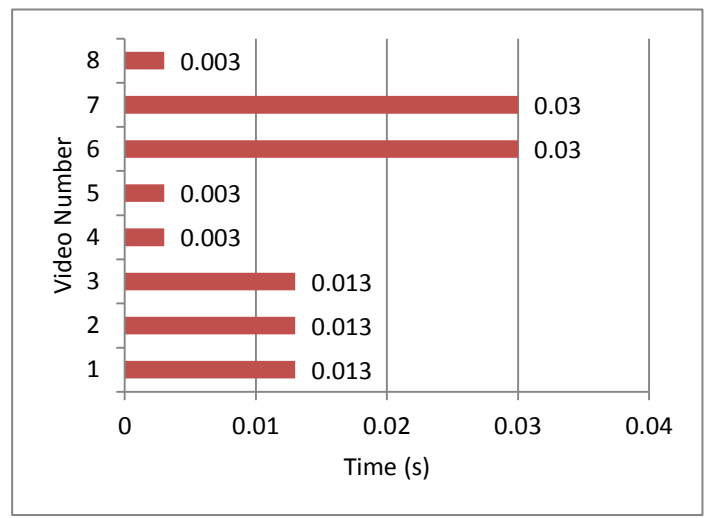

Fig.15. Time of working state detection for different video datasets

The result of object region detection, feature extraction, and EWMA data processing, data labelling, and classification in the algorithm flow of this paper are shown in Fig. 16. Fig. 16(c) shows that the detection of the object region by our method is accurate. Fig. 16(d) shows that features extracted based on the RF-KLT algorithm have very distinct discrimination and stability under various scenarios. After the K-means clustering method labels the data, the support vectors of the SVM algorithm are all in the process of the working state change, that is, the feature vector from the beginning of the spray to reach the maximum region. There is a clear separation between the two classes of constructed dataset in Fig. 16(e).

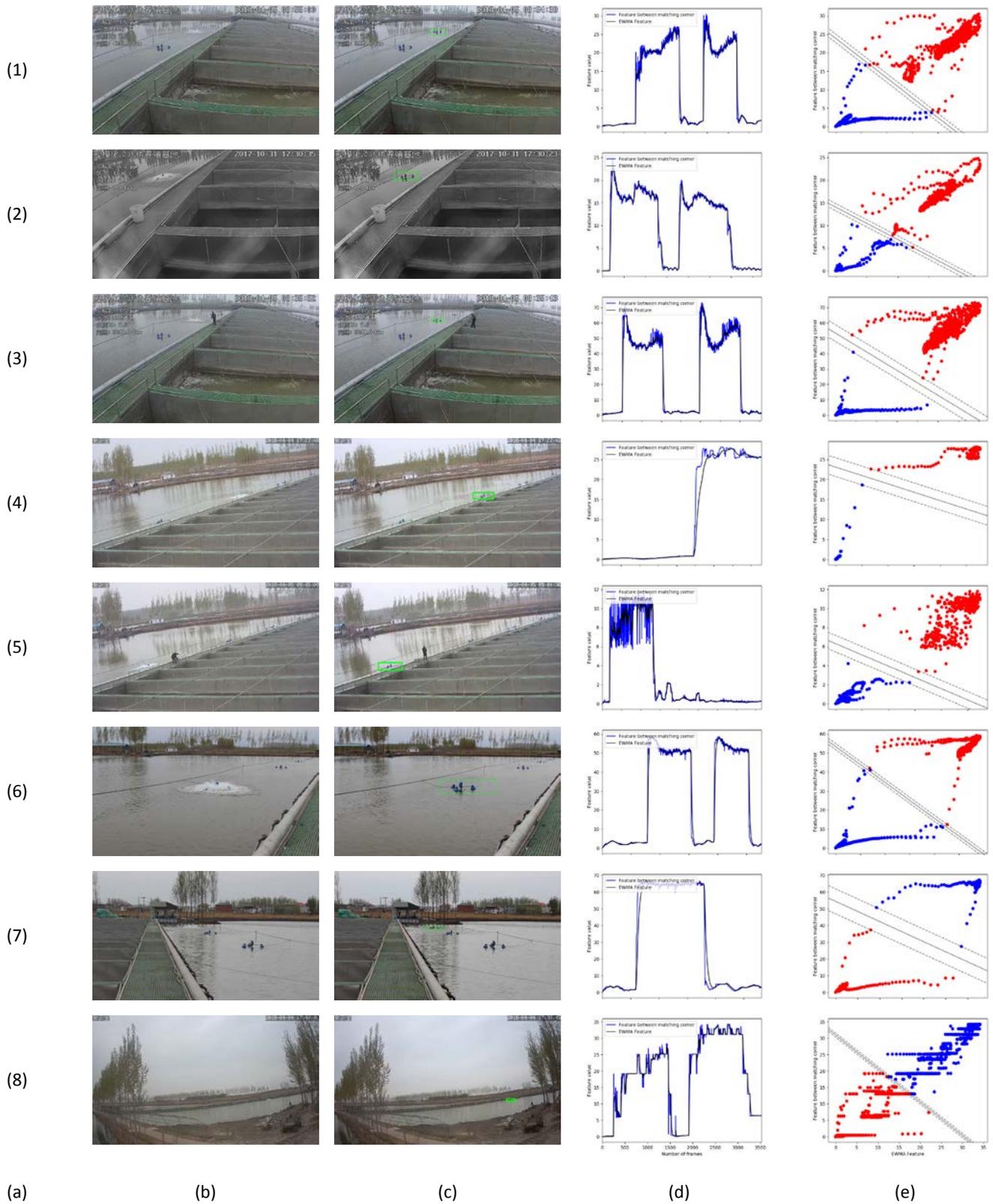

(a) (b) (c) (d) (e)

Fig. 16. The result of key algorithm flow in video dataset: (a) is the video number, (b) is a frame of the opening state of aerators in video dataset. (c) is the result of the object region detection. (d) is the feature curve based on the RF-KLT algorithm and the data after EWMA processing, and (e) is the result of dataset construction, data annotation and classification.

However, in a very small object region, because the number of detected corner points is small and the area of the spray region is small, resulting in a smaller corner distance feature. Therefore, the discrimination of datasets is not obvious. Fig. 17 shows the some results in algorithm flow for P1-P4, which are video datasets with artificially augmented interference. The results show that the influence of brightness is greater than that of noise, and the randomly changed brightness does not affect the experimental results. A non-skipping increase in brightness does not affect the detection results, and even better situations may occur. But when the amount of brightness change is extreme and constant, the instability of the feature data is increased. This is because when the brightness of the object region changes sharply, the corner points detection based on the feature value of gray variation will change, resulting in a change of the distance features between matching corners.

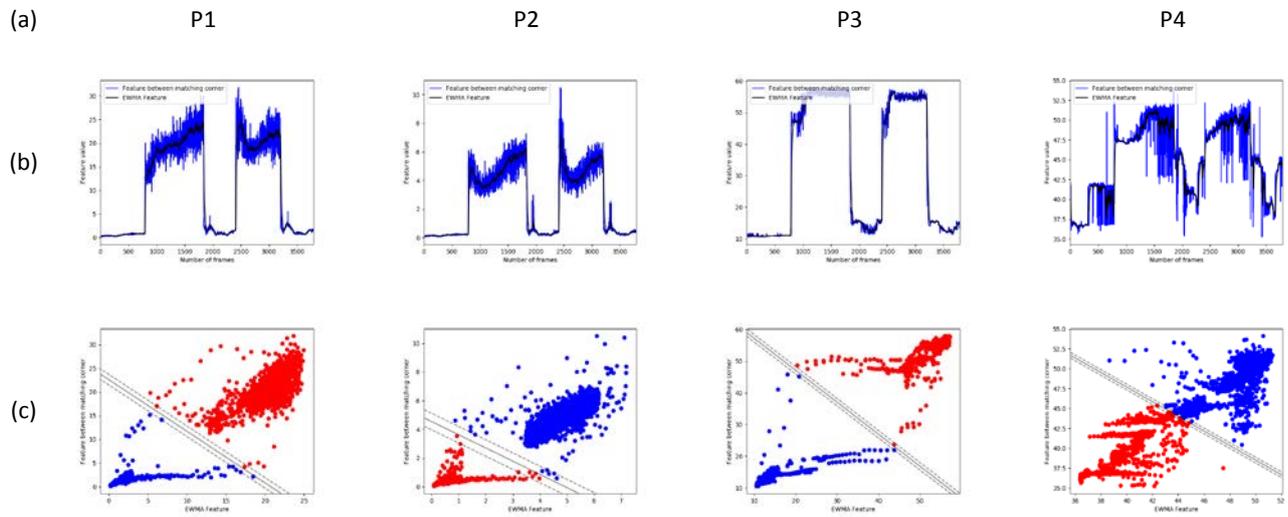

Fig. 17. The result of key algorithm flow in the artificially augmented dataset: (a) is the video number, (b) is the feature curve based on the RF-KLT algorithm and the data after EWMA processing, and (c) is the result of dataset construction, data annotation and classification. The artificially augmented video dataset is based on video 1, so the object region is the same as the object region in video 1.

*4.4. Suggestions in application*

Based on practical applications and the problem of intense change in brightness, the authors suggest: (1) In order to avoid the influence of drastic change in the light intensity in different seasons or time periods, light sensors can be used to measure the dramatic change in light and the classifier model is updated accordingly. The authors suggest updating the classifier when the steady increase of gray value reaches 80. (2) Feature is extracted every 5 seconds, and the working state is classified by combining multiple consecutive results in scenes of light strongly changed. (3) The working state detection algorithm in this study can use a single surveillance camera to monitor multiple aerator regions at the same time, but the object region detection module needs to be revised.

## 5. Conclusions and future works

In this paper, we present a real-time expert system with existing surveillance cameras for anomaly detection of aerators, which consists of object region detection and working state detection. In the object region detection module, we propose a regional selection method for detecting small object region from complex background based on region proposal idea (Girshick et al.,

2014). This method includes three steps, i.e., maximum contour region detection, candidate region detection, and target region detection. In the working state detection module, we propose the RF-KLT algorithm for robust motion feature extraction in fixed regions, which is simply based on the reference idea. The RF-KLT algorithm extends the applicability of the conventional optical flow method and breaks the limitation of adjacent frames. Moreover, we present a dimension reduction method of time series for extracting the numerical distribution characteristics. Finally, we establish an SVM classifier for work state judgment of aerators. The experimental results show that the accuracy of object region detection is 100%, the average accuracy of working state detection is 99.9%, and the detection speed is between 77-333 FPS according to different surveillance cameras. The proposed expert system can achieve real-time, zero-cost and accurate anomaly detection of aerators in complex and different application scenarios, which can be easily applied as part of the intelligent agriculture and agricultural expert system.

In future, we aim to improve this expert system so that it can be monitored for twenty-four hours, because this study was based on video dataset during the day and could not be monitored at night unless lighting equipment with high power was used, but it is a waste of energy. Accordingly, we expect to use surveillance cameras with infrared night vision equipment or other night vision equipment for imaging and analysis. Besides, the reference idea shows that the matching method of feature points with a certain reference frame can be used to detect motion state in videos. Therefore, different feature detection algorithms (e.g., Harris, SIFT, ORB, etc.) can be attempted. Finally, the proposed method for detecting small regions with fixed motion features from complex scenes, and the fast modeling method for numerical distribution of time series can be both applied to more practical scenarios.

**Acknowledgments**

This research is supported by the Science & Technology Program of Beijing "Research and Demonstration of technologies equipment capable of intelligent control for large-scale healthy cultivation of freshwater fish"(No. Z171100001517016), and the Shandong Province key Research & Development Program "Research and Demonstration of accurate monitoring and controlling technologies for environment of vegetable in facility"(No.2017CXGC0201).